\documentclass{article}

\usepackage[english]{babel}

\usepackage[letterpaper,top=2cm,bottom=2cm,left=3cm,right=3cm,marginparwidth=1.75cm]{geometry}

\usepackage{amsmath}
\usepackage{graphicx}
\usepackage{cite}
\usepackage{authblk}
\usepackage[colorlinks=true, allcolors=blue]{hyperref}

\title{OpenGo: An OpenClaw-Based  Robotic Dog with Real-Time Skill Switching}
\author{Hanbing Li, Xuewei Cao, Zhiwen Zeng, Yuhan Wu, Yanyong Zhang, Yan Xia\thanks{Corresponding author}}
\affil{University of Science and Technology of China}

\begin{document}
\maketitle

\begin{abstract}
Adaptation to complex tasks and multiple scenarios remains a significant challenge for a single robot agent. The ability to acquire organize, and switch between a wide range of skills in real time, particularly in dynamic environments, has become a fundamental requirement for embodied intelligence. We introduce OpenGo, an OpenClaw-powered embodied robotic dog capable of switching skills in real time according to the scene and task instructions. Specifically, the agent is equipped with (1) a customizable skill library with easy skill import and autonomous skill validation, (2) a dispatcher that selects and invokes different skills according to task prompts or language instructions, and (3) a self-learning framework that fine-tunes skills based on task completion and human feedback. We deploy the agent in Unitree's Go2 robotic dog and validate its capabilities in self-checking and switching of skills autonomously. In addition, by integrating Feishu-platform communication, we enable natural-language guidance and human feedback, allowing inexperienced users to control the robotic dog through simple instructions.
\end{abstract}

\section{Introduction}

Robust quadruped robot tasks in diverse environments remain a fundamental challenge. Robots must simultaneously deal with rapidly changing environments, indoor or outdoor spaces, sensory degradation, and safety behavioral limitations. Even strong reinforcement-learning-based quadruped controllers that show zero-shot transfer on rough outdoor terrain still face significant difficulty in harder conditions where perception is unreliable, such as snow, plants, water, fog, reflective surfaces, or partial occlusion.\cite{lee2020challengingterrain} In practice, this is why recent high-performance navigation systems for quadruped robots still incorporate multiple obstacle or terrain-aware skills, such as walking, jumping, going up and down stairs, climbing, and crouching, rather than relying on a single general strategy.\cite{miki2022perceptive,tang2023saytap} This trend suggests a key limitation for embodied agents: while a general skill may work in nominal settings, its re-usability often degrades sharply in extreme environments, where success depends on explicit awareness of scene geometry, safety margins, and the robot’s executable capability set.\cite{hoeller2024anymalparkour,zhuang2023robotparkour}

\begin{figure}
    \centering
    \includegraphics[width=0.9\linewidth]{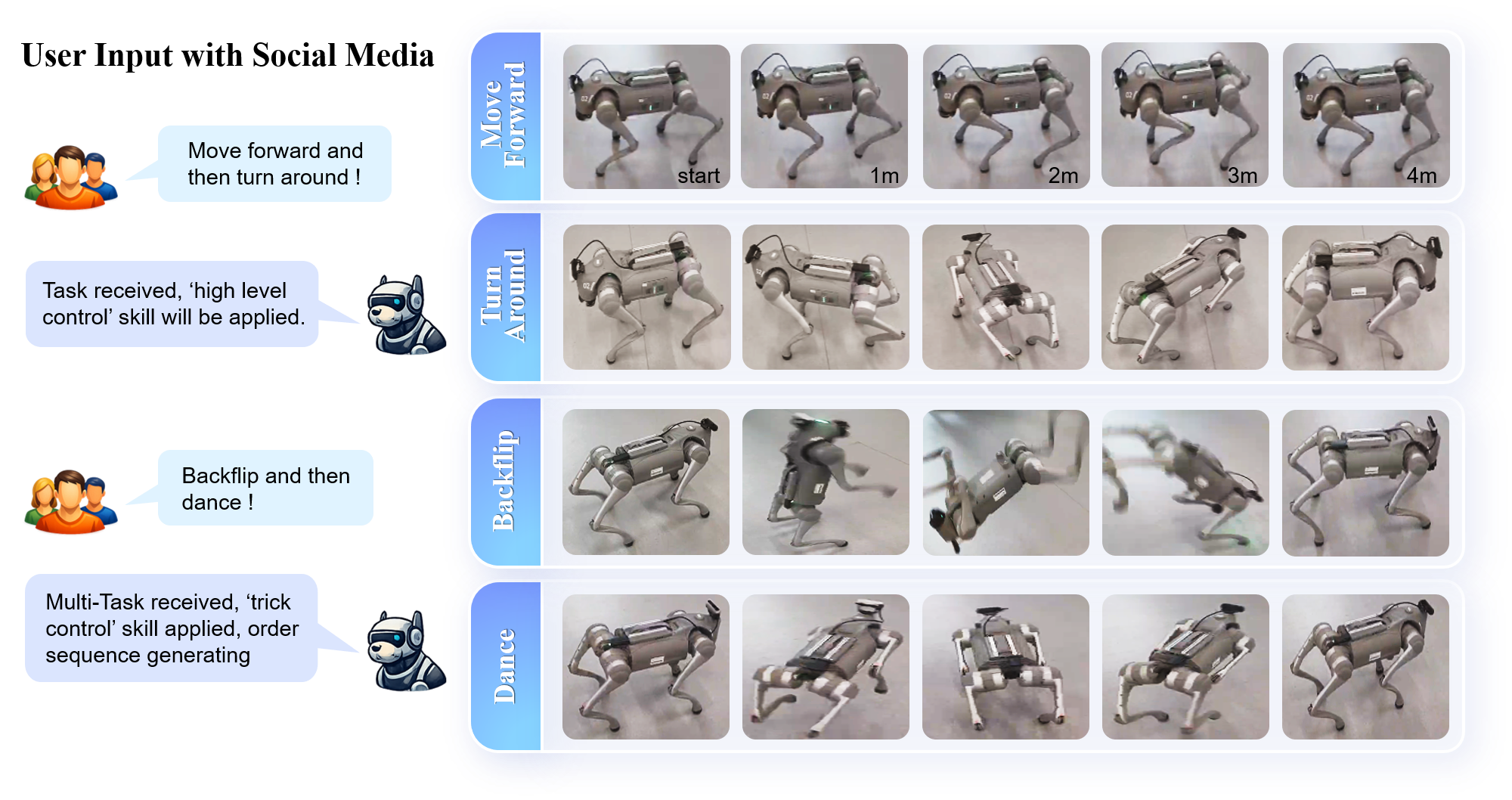}
    \caption{Demonstration of LLM-driven skill execution and composition on the quadruped robot. 
    Given natural language instructions (left), the system interprets user order and transmit it to Unitree Go2. From top to bottom, the commands are: Move Forward, Turn Around, Backflip, and Dance.}
    \label{fig:1}
\end{figure}

At the same time, embodied intelligence requires a robot to unite perception, semantic grounding, planning, and action under high-level human intent. At the same time, embodied intelligence requires a robot to pair perception, semantic grounding, planning, and action under high-level human intent. Recent large pretrained models have made this increasingly feasible by transferring semantic knowledge learned from internet-scale language and vision corpora into robot decision-making.\cite{ding2023quarvla} PaLM-E shows that a single embodied multimodal model can transfer knowledge from visual-language domains to embodied reasoning and robot planning.\cite{driess2023palme} RT-2 further demonstrates that vision-language-action models can transfer web knowledge directly into robotic control, improving generalization and semantic reasoning.\cite{zitkovich2023rt2} SayCan shows that large language models can be grounded by a library of pretrained skills and affordance/value estimates, enabling long-horizon task execution on real robots.\cite{ahn2022saycan,ouyang2024longhorizon} In summary, these results strongly support the feasibility of using LLMs as high-level decision makers in embodied systems, provided that semantic reasoning is tightly coupled to embodiment-aware action grounding.

In simulated or reinforcement-learning environments, LLM-driven agents have already shown impressive decision-making capacity. Voyager uses an automatic curriculum, a code-based skill library, and iterative prompting to achieve open-ended lifelong skill acquisition in Minecraft. Eureka uses GPT-4 to synthesize reward code and outperforms human-engineered rewards across a broad suite of reinforcement-learning tasks.\cite{ma2024eureka} More generally, prior work has shown that language models can generate executable plans for embodied agents and improve these plans with environment feedback.\cite{huang2022zeroshot} However, transferring this paradigm from simulation to physical robots is substantially harder. As noted by SayCan, an unconstrained language model may generate a plan that is linguistically reasonable yet infeasible for a specific robot in a specific environment.\cite{ahn2022saycan} In robotics, this is the practical form of hallucination: invalid skill calls, unsafe parameter assignments, or behavior that ignores embodiment and scene constraints. \cite{huang2023innermonologue}This is precisely why direct end-to-end reliance on free-form LLM outputs remains difficult in real robot deployment.

To address this issue, we present OpenGo, an OpenClaw-powered embodied robotic dog pipeline in which controllability is anchored in the skill layer rather than in unconstrained language generation. OpenGo maintains a customization skill library, and the role of the LLM is deliberately restricted to two high-level functions: selecting the most appropriate skill for the current scene and task, and setting a bounded set of skill hyperparameter according to task prompts, natural-language guidance, and feedback. By design, this reduces the impact of hallucination, because the final behavior must be instantiated through predefined, robot-compatible, and self-checkable skills. In this way, OpenGo preserves the semantic flexibility of language-guided decision making while improving robustness, interpret-ability. This pipeline has been examined on  a real Unitree Go2 platform.

\section{Related work}

\subsection{LLM-driven decision making}

Recent work has shown that large language models can serve as effective high-level controllers for embodied systems. Code as Policies uses language models to generate executable robot policy code that calls perception modules and control units, enabling reactive and waypoint-based behaviors on real robots.\cite{liang2022codeaspolicies} ProgPrompt further improves executability by constraining the language model with program-like specifications of available actions and objects, so that task plans remain consistent with the robot’s capabilities and the current scene.\cite{singh2023progprompt} VIMA demonstrates that a transformer-based agent can solve a broad family of manipulation tasks from multimodal prompts, suggesting that language-like prompting can unify task specification across text, images, and demonstrations.\cite{jiang2023vima} Compared with direct low-level action generation, these approaches collectively indicate that LLMs are most reliable when they operate over structured abstractions—such as APIs, plans, or skills—rather than raw motor commands.

\subsection{Skill libraries and embodied agents.}

Within embodied intelligence, the notions of skill library and agent have become increasingly important as a way to bridge semantic reasoning and physical execution. A skill library represents behavior as a reusable set of callable modules, often parameterized and composable, while the agent is responsible for selecting, sequencing, and adapting those skills according to task context. This pipeline is attractive because it improves modularity, interpretability, reusability, and controllability compared to monolithic end-to-end policies. 

\begin{figure}
    \centering
    \includegraphics[width=0.75\linewidth]{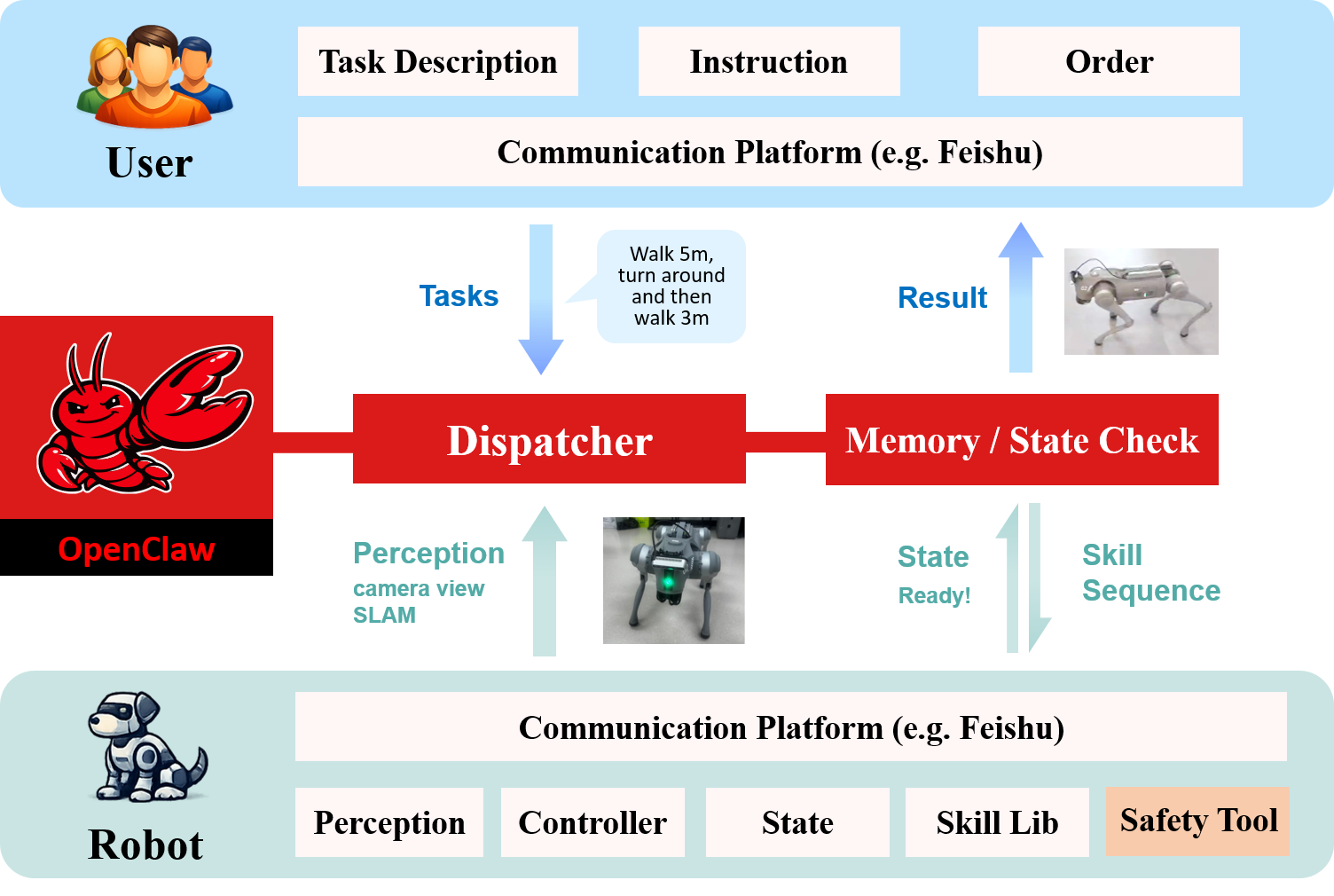}
    \caption{Overview of the OpenGo framework. OpenGo is built upon OpenClaw and organized around two core modules, namely the \textbf{Dispatcher} and \textbf{Memory/State Check}. Through a communication platform (e.g., Feishu), human users provide task descriptions, instructions, and execution orders to the system. The Dispatcher selects appropriate skills from the robot-side skill library, while the Memory/State Check module monitors execution status and feeds state information back to support closed-loop decision making. On the robot side, the framework interfaces with perception, controller, state estimation, skill library, and an internal \textbf{Safety Tool}, which serves as an emergency-stop trigger when the robot enters a dangerous state, thereby enabling controllable and robust skill execution on the quadruped platform.}
    \label{fig:2}
\end{figure}

Voyager is a representative example: it maintains a growing library of executable skills and reuses them to solve unseen tasks in an open-ended environment.\cite{wang2023voyager} More recently, \textbf{BOSS (Bootstrap Your Own Skills)} learns to solve new long-horizon tasks by expanding a language-specified skill library with minimal supervision and then reusing those acquired skills in new environments.\cite{tziafas2024lrll} Together, these works suggest that a skill-library-based embodied agent provides a practical middle ground between symbolic planning and low-level control, especially when the system must remain extensible, interpretable, and robust under changing task demands. These approaches improve semantic controllability, but they also reveal a common limitation: high-level reasoning remains dependent on either carefully designed symbolic representations, fixed low-level skill set, or a constrained scene., which leads to open-loop plans fail when the robot enters unexpected states.

\subsection{High-level control for quadruped robots. }

Recent quadruped research has explored several routes toward high-level motion control, each with distinct strengths and limitations. One line of work focuses on perception-aware locomotion: for example, robust perceptive locomotion integrates external and internal sensing to achieve fast and stable traversal in challenging outdoor environments, but its scope remains largely locomotion-centric rather than instruction-centric.\cite{miki2022perceptive} Another way uses hierarchical multi-skill control. ANYmal Parkour trains multiple specialized locomotion skills—such as walking, jumping, climbing, and crouching—and then employs a high-level policy to switch among them according to the scenario, having strong agility but also relying on a predefined repertoire of obstacle-aware skills.\cite{hoeller2024anymalparkour}  Robot Parkour Learning distills diverse parkour behaviors into a single end-to-end vision-based policy, reducing hierarchy design and enabling autonomous skill selection, but the system still faces practical issues such as perception delay and extreme motor demand.\cite{zhuang2023robotparkour}

\section{Method}

We propose \textbf{OpenGo}, an embodied quadruped agent built on top of \textbf{OpenClaw} for robust task execution across diverse scenarios. As shown in Fig. 1, OpenGo is designed around three tightly coupled components: a \textbf{customizable skill library}, an \textbf{LLM-based dispatcher}, and a \textbf{self-learning framework} driven by execution outcomes and human feedback. The key design principle of OpenGo is to preserve the semantic flexibility of large language models while constraining their decision space to a set of validated robot skills. Instead of allowing the LLM to directly generate low-level robot actions, OpenGo restricts it to selecting skills from a structured library and assigning bounded hyperparameters under task and scene constraints. This design improves safety, controllability, and deployability on a real quadruped platform. 

Given a task description \(T\), a human instruction \(I\), and a perceived scenario \(S\), the system dispatcher generates a skill execution plan 
\(\prod {(k_1,\theta_1),(k_2,\theta_2),...,(k_n,\theta_n)}\)
where \(k_i\)denotes a callable skill from the library and \(\theta_i\)denotes its task-dependent parameters. During execution, OpenGo monitors robot states and collects completion signals, error logs, and user feedback, which are then used to improve future dispatching and skill usage. In the following, we describe the three core components of the system. 

\begin{figure}
    \centering
    \includegraphics[width=1\linewidth]{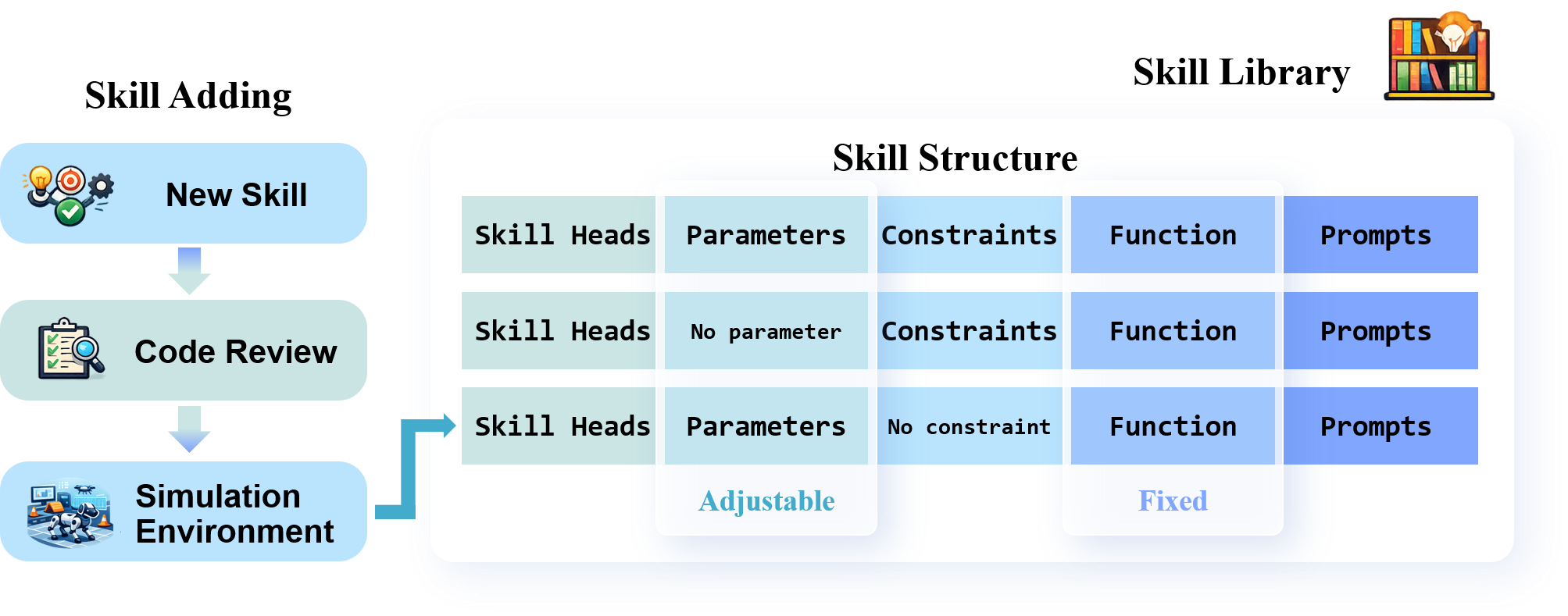}
    \caption{\textbf{Skill library design in OpenGo.} New skills are incorporated through code review and simulation-based validation before entering the skill library. Each skill is organized with structured fields, including \textbf{skill heads}, \textbf{parameters}, \textbf{constraints}, \textbf{function}, and \textbf{prompts}. The \textbf{parameters} are adjustable for task adaptation, whereas the \textbf{function} remains fixed for stable and controllable execution. 
    \label{fig:3}}
\end{figure}

\subsection{Customizable Skill Library}

The skill library is the foundation of OpenGo. Its purpose is to provide a \textbf{structured, reusable, and controllable action space} for the quadruped robot. Each skill corresponds to an executable robot capability, such as locomotion, posture adjustment, obstacle negotiation, navigation, or task-specific interaction. Rather than treating all robot behavior as an end-to-end policy, OpenGo decomposes execution into a set of callable skill modules that can be explicitly managed, validated, and reused. 

As illustrated in Fig. 2, each skill is represented by a structured template containing five fields: \textbf{skill head}, \textbf{parameters}, \textbf{constraints}, \textbf{function}, and \textbf{prompts}. The \textbf{skill head} defines the skill identifier and its semantic label. The \textbf{parameters} specify adjustable hyperparameters for task adaptation, such as target distance, speed, duration, turning angle, or environment-dependent thresholds. The \textbf{constraints} define preconditions and safety boundaries for execution, such as terrain requirements, valid robot states, actuator limits, or forbidden transitions. The \textbf{function} contains the actual implementation of the skill and is kept fixed once validated, ensuring consistent and robot-compatible execution. The \textbf{prompts} provide textual descriptions or usage instructions that help the dispatcher understand when and how the skill should be invoked. 

A central feature of our library design is the separation between \textbf{fixed execution function} and \textbf{adjustable task configuration}. In particular, the \textbf{function} is immutable during online execution, while the \textbf{parameters} remain tunable within predefined bounds. This separation prevents the LLM from modifying low-level control behavior directly, while still allowing the system to adapt to task requirements at the semantic level. As a result, OpenGo maintains a controllable interface between high-level language reasoning and low-level robot execution. 

The skill library is also customizable and extensible. When a new skill is introduced, it does not enter the library immediately. Instead, it passes through a multi-stage import process consisting of: (1) \textbf{code review}, to verify interface compatibility, parameter definitions, and safety assumptions; and (2) \textbf{simulation-based validation}, to test executability and robustness before deployment. Only skills that pass these checks are admitted into the library. This process allows OpenGo to continuously expand its capability set while preserving reliability. In this sense, the library functions not only as an action repository, but also as a safety- and consistency-enforcing abstraction layer for embodied decision making. 

\subsection{The Dispatcher }

The dispatcher is responsible for translating high-level human intent into an executable skill sequence. As shown in Fig. 3, it receives three main categories of input: (1) a \textbf{task description}, which specifies the overall goal; (2) \textbf{human instructions}, which may refine preferences, order, or operating details; and (3) the current \textbf{scenario state}, inferred from the perception and classification modules. Based on these inputs, the dispatcher calls the LLM to generate a structured execution plan over the skill library.

Unlike unconstrained language-based robot control, the OpenGo dispatcher does not permit the LLM to output arbitrary actions. Instead, the LLM operates only over a predefined and validated skill set. Concretely, its role is restricted to two functions:

\begin{enumerate}
    \item \textbf{skill selection}, i.e., choosing which skill or sequence of skills should be executed under the current task and scenario; 
    \item \textbf{parameter assignment}, i.e., setting bounded skill hyperparameters according to the task requirement and environmental context. 
\end{enumerate}
Formally, given task \(T\), instruction \(I\), and scene \(S\), the dispatcher computes 
\[\prod =f_{LLM}(T,I,S,K)\]

where \(K\)denotes the skill library and  \(\prod \)is an ordered sequence of skill-parameter pairs. The output is therefore not a motor command stream, but a symbolic-to-structured plan grounded in validated robot skills. 

To further reduce hallucination and infeasible outputs, we impose several constraints on dispatching. First, skill candidates are filtered by skill constraints and current perception scenarios before being presented to the LLM. Second, each parameter is restricted to a valid range defined in the corresponding skill template. Third, all selected skills are checked against current execution context through the Memory/State Check module. This module maintains recent execution history, monitors whether the robot is in a valid state for the next step, and provides failure or completion feedback for dynamic replanning. 

\subsection{Self-Learning Framework }

The third component of OpenGo is a self-learning framework that improves skill usage over time through execution outcomes and human interaction. The goal of this module is not to let the robot freely rewrite its own control code online, but to refine the way skills are selected, parameterized, and reused based on observed success and failure.

During task execution, OpenGo continuously records three types of feedback: (1) \textbf{task completion signals}, indicating whether the goal or subgoal has been achieved; (2) \textbf{error logs}, including failed skill calls, invalid transitions, interrupted execution, or abnormal robot states; and (3) \textbf{human feedback}, which is delivered through the communication platform and may include corrections, preferences, or explicit guidance. This feedback is routed back to the dispatcher and memory module, forming a closed-loop execution pipeline.

The self-learning framework uses these signals at two levels. At the \textbf{dispatch level}, it updates the preference for skill selection under specific scenarios. For example, if one skill consistently fails under a given terrain class while an alternative skill succeeds, the system increases the priority of the successful skill for similar future conditions. At the \textbf{parameter level}, it adjusts default hyperparameter settings within the valid range defined by the skill template. For instance, step length, turning radius, or execution duration may be tuned based on repeated success or failure patterns. In this way, OpenGo gradually improves task completion without violating the structural constraints of the skill library. 

Human feedback plays an especially important role when task ambiguity cannot be resolved from perception alone. Since the system is connected to a communication platform such as \textbf{Feishu}, non-expert users can provide natural-language instructions during or after execution, such as requesting a different behavior, confirming completion, or correcting an undesirable plan. These user signals are converted into additional supervisory information for future dispatching. This allows OpenGo to incorporate human preferences without requiring direct teleoperation or manual policy engineering. 

Overall, the self-learning framework enables OpenGo to move beyond static task execution. By combining completion signals, execution logs, and human feedback, the system can iteratively improve its decision quality while preserving a safe and interpretable skill structure. This makes OpenGo particularly suitable for practical embodied deployment, where robot behavior must remain both adaptable and controllable over long-term use. 

\begin{figure}
    \centering
    \includegraphics[width=1\linewidth]{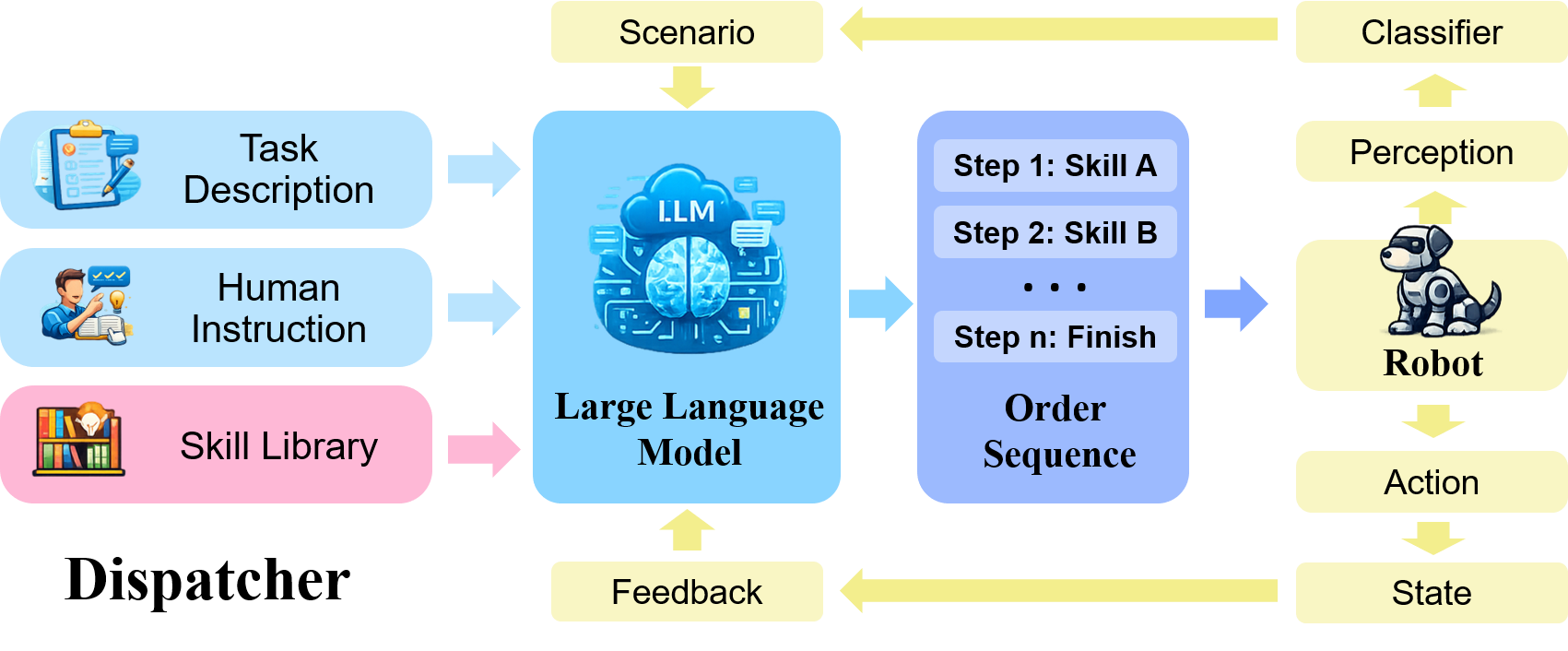}
    \caption{\textbf{Dispatch mechanism of OpenGo.} Task descriptions, human instructions, and scene information inferred from perception are jointly provided to the LLM-based dispatcher. The LLM selects appropriate skills from the skill library and organizes them into a step-by-step execution sequence. Execution feedback, including error logs and finished signals, is then returned to the dispatcher, enabling dynamic replanning and closed-loop skill scheduling. }
      \label{fig:4}
\end{figure}
 
\section{Experiments}

In this section, we evaluate the proposed OpenGo framework on a real quadruped platform and validate its feasibility in skill generation, deployment, human–robot interaction, and system-level latency performance.

\subsection{System Deployment }

We deploy OpenGo on the Unitree Go2 quadruped robot, integrating the OpenClaw framework with the robot’s onboard control system. The overall pipeline follows a three-stage process: skill generation, simulation validation, and real-world deployment.

First, the large language model is used to generate basic action control units, which serve as candidate skills for the robot. The generated code then undergoes a code review process, followed by automatic validation in a simulation environment to ensure correctness, safety, and compatibility before deployment. These control units are defined in a structured format compatible with the OpenGo skill library, including parameters and execution interfaces. Second, the skills generated are tested in a simulation environment to verify their correctness, stability, and compatibility with the robot’s control constraints. Only skills that pass this validation stage are incorporated into the skill library. Finally, the validated skills are deployed on the real Go2 robot, where they are executed through the dispatcher under task and scenario conditions.

This pipeline ensures that all executable behaviors used on the physical robot have undergone prior verification, reducing the risk of unsafe or infeasible actions during real-world operation. We provide the dispatcher promotion codes in the appendix.

\begin{figure}
    \centering
    \includegraphics[width=0.5\linewidth]{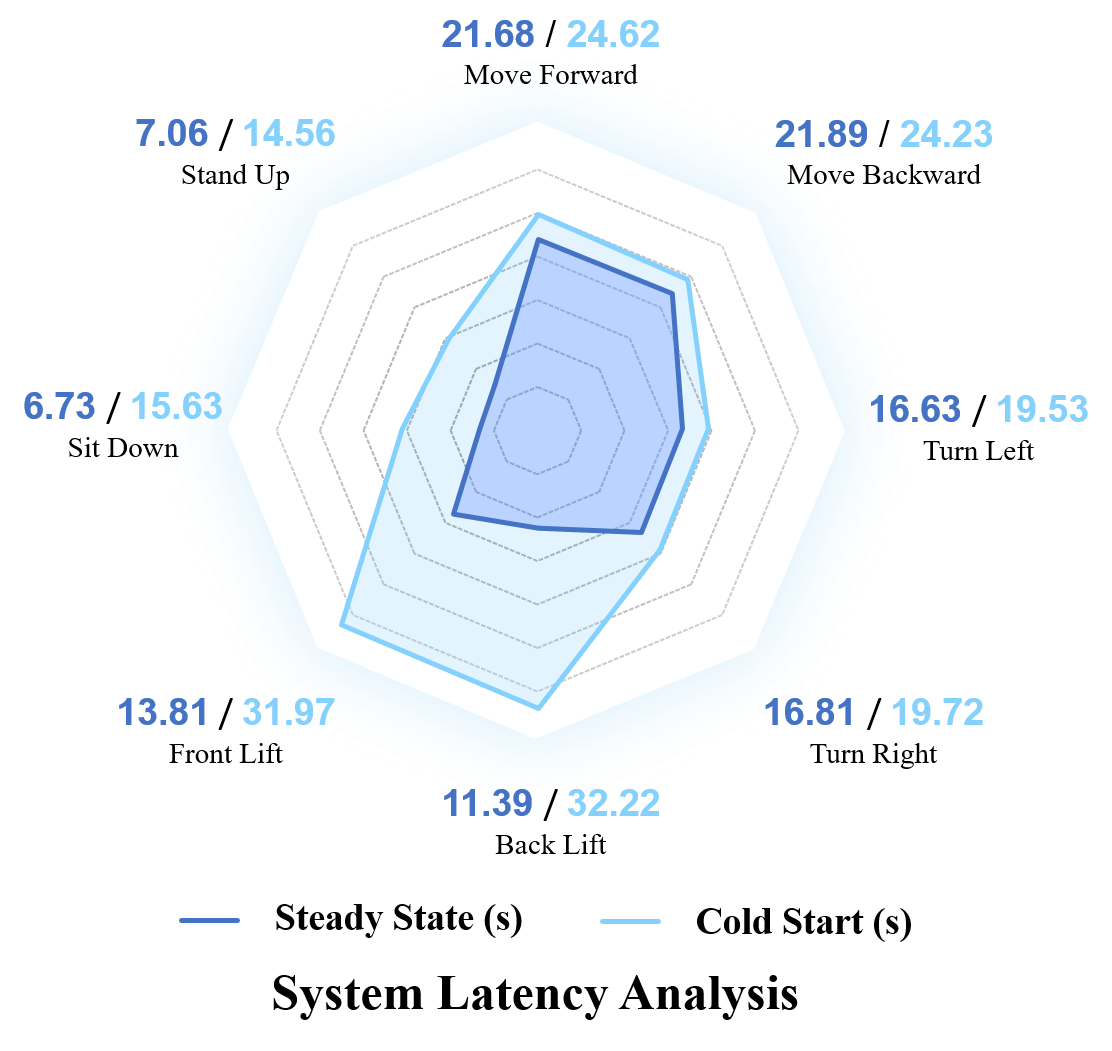}
    \caption{System Latency Analysis of Single-Skill. Experiments are conducted on a real-world Unitree Go2 platform. Each action is executed 10 times, and the response time is measured from the moment the user issues the instruction to the initiation of the robot’s execution.}
    \label{fig:5}
\end{figure}

\subsection{Human–Robot Interaction via Communication Platform}

To enable intuitive and accessible interaction, we integrate the system with the Feishu communication platform, allowing users to send task descriptions, instructions, and commands directly to the robot. Through this interface, users can interact with OpenGo in natural language without requiring prior robotics expertise.

The communication platform serves as a bridge between human input and the LLM-based dispatcher. User instructions are transmitted to the system, interpreted by the LLM, and translated into structured skill sequences. During execution, the robot returns feedback signals, including execution status and error logs, back to the user through the same platform. This bidirectional communication enables a closed-loop interaction process, where users can monitor progress and provide additional guidance if necessary.

In practice, this setup significantly lowers the barrier for controlling quadruped robots, making it possible for inexperienced users to perform task-level control through simple language instructions.

\subsection{System Latency Analysis}

To quantitatively evaluate the responsiveness of the OpenGo framework, we conduct a system-level latency analysis, decomposing the response time into two scenarios: (1) single-skill execution and (2) multi-skill composition.

\textbf{Single-Skill Latency.}  
We measure the response time from the moment a user instruction is issued to the initiation of the corresponding robot action. As illustrated in \ref{fig:5}, the latency shows a clear distinction between the first invocation and subsequent executions of the same skill. Specifically, initial execution incurs a significantly higher delay due to skill loading and initialization overhead, including LLM parsing, skill retrieval, and parameter instantiation. In contrast, subsequent executions benefit from cached or preloaded skill representations, resulting in reduced response times.

Furthermore, we observe that latency increases with the number of input parameters associated with a skill. This is primarily due to the increased complexity in parsing, validation, and binding of parameters within the LLM-based dispatcher, as well as additional overhead in the underlying control interface.

\begin{figure}
    \centering
    \includegraphics[width=0.9\linewidth]{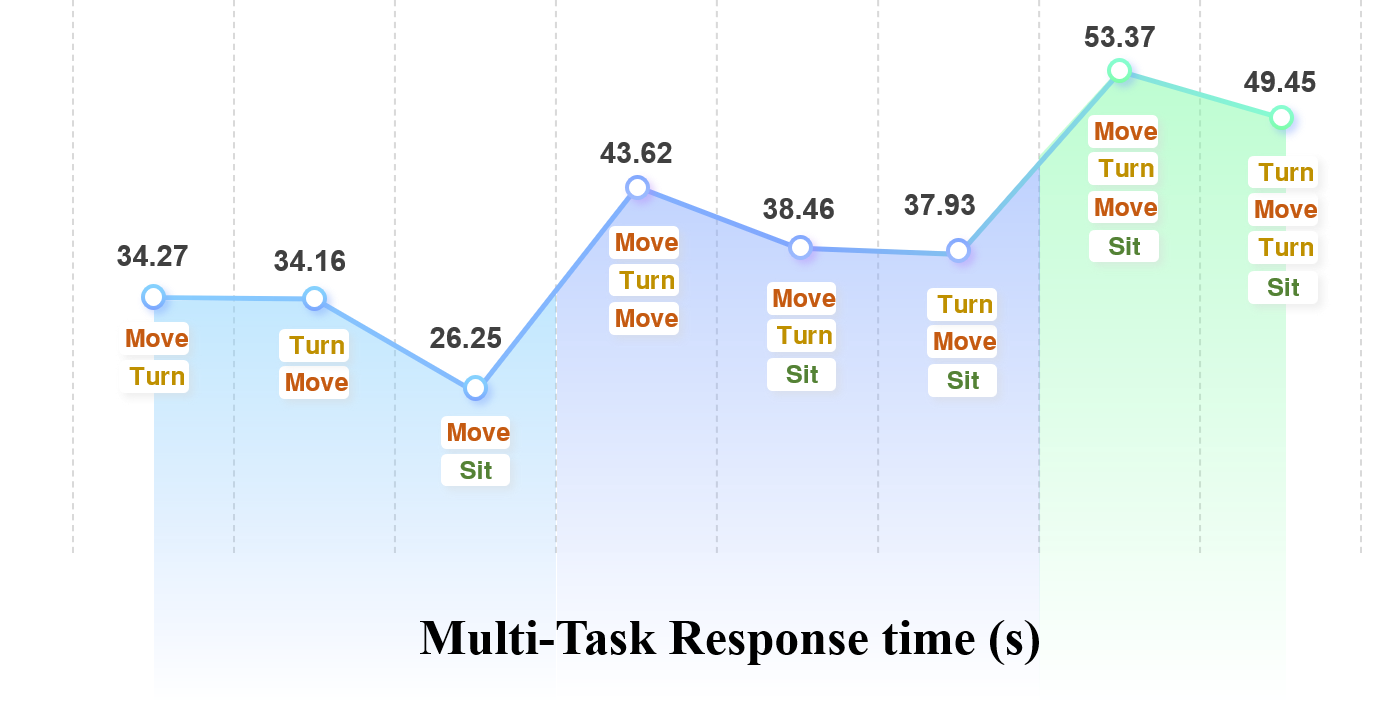}
    \caption{System Latency Analysis of Multi-Task. We analyze the response time for compositions of 2–4 individual instructions. The order of certain instructions is varied, and each combination is executed five times after the system reaches a stable state.}
    \label{fig:6}
\end{figure}

\textbf{Multi-Skill Composition Latency.}  
We also evaluate scenarios that involve sequences of multiple skills. As shown in \ref{fig:6}, the total response time does not scale linearly with the number of skills, indicating that the system introduces additional coordination overhead beyond simple accumulation of individual execution times. This includes inter-skill scheduling, state transitions, and dependency resolution.

Despite this, a general trend of increasing latency is observed as the number of composed skills grows. Similarly to the single-skill case, the complexity of the parameter of each instruction also plays a critical role in determining the overall response time. Tasks requiring more parameters or more complex configurations tend to incur higher latency.

\textbf{Discussion.}  
These findings highlight two key system characteristics: (1) the presence of a “cold-start” latency associated with skill initialization, and (2) the impact of parameter complexity on execution efficiency. From a system design perspective, these results suggest that performance can be improved through skill preloading, caching mechanisms, and more efficient parameter parsing strategies. Additionally, optimizing the execution pipeline for multi-skill coordination is critical for enabling smooth long-horizon task execution.

\subsubsection{More Complex Skill Composition }

In the current implementation, we focus on validating the pipeline using basic action control units, demonstrating the feasibility of LLM-guided skill generation and execution. These basic skills provide the foundation for more advanced behaviors.

Building on this framework, we plan to extend the skill library with more complex and structured control modules, such as obstacle avoidance and navigation modules. These higher-level skills will further enhance the robot’s ability to operate in dynamic and unstructured environments, enabling more robust long-horizon task execution.

The modular design of OpenGo allows such extensions to be incorporated naturally through the existing skill import and validation pipeline, without requiring changes to the overall system architecture.

\section{Limitations}

Despite the effectiveness of OpenGo in enabling structured and controllable embodied decision making, several limitations remain in the current system.

First, the response latency of the large language model introduces a noticeable delay in the decision pipeline. Specifically, the process from natural language understanding to skill retrieval and final generation of the execution sequence requires a non-negligible initialization time. This latency limits the responsiveness of the system, particularly in time-sensitive or dynamic environments where rapid decision making is critical. The issue becomes more pronounced when tasks require frequent replanning or iterative interaction with the user.

Second, the execution latency within OpenClaw affects the temporal continuity of robot behavior. Each skill invocation incurs a delay, and transitions between consecutive skills introduce additional gaps. As a result, the overall motion of the quadruped robot may appear fragmented rather than smooth and continuous. This limitation is especially evident in tasks that require fine-grained coordination or fluid motion, such as dynamic navigation or obstacle negotiation.

\section{Summary}

In this work, we present OpenGo, an embodied quadruped agent that integrates large language model reasoning with a structured and controllable skill-based execution framework. By introducing a customizable skill library, an LLM-based dispatcher, and a self-learning framework grounded in execution feedback and human interaction, OpenGo provides a practical approach to bridging high-level semantic understanding and low-level robot control.

A key contribution of this work lies in constraining the role of the LLM to skill selection and parameter tuning, rather than direct generation of low-level actions. This design significantly improves the safety, interpretability, and deployability of LLM-driven embodied systems. Through deployment on the Unitree Go2 quadruped robot, we demonstrate that the proposed framework can support real-world task execution, skill switching, and natural-language-based human–robot interaction via a communication platform.

While current limitations remain in system latency and execution continuity, the overall framework establishes a scalable foundation for extending embodied intelligence systems toward more complex, dynamic, and interactive environments. Future improvements in skill composition, real-time control, and adaptive learning are expected to further enhance the robustness and generalization capability of the system.

In summary, OpenGo highlights a feasible and effective pathway for combining large language models with structured robotic skill systems, moving toward more reliable and accessible embodied AI.

\newpage

\bibliographystyle{unsrt}
\bibliography{references}

\end{document}